\documentclass[a4paper,conference]{IEEEtran}
\ifCLASSINFOpdf
\else
\fi
\usepackage[tight]{subfigure}
\usepackage{graphicx}
\usepackage{amsmath}
\usepackage{amssymb}
\usepackage{url}

\begin{document}
%
\title{Domain Siamese CNNs for Sparse Multispectral Disparity Estimation}

\author{
\IEEEauthorblockN{David-Alexandre Beaupre and Guillaume-Alexandre Bilodeau}
\IEEEauthorblockA{LITIV lab., Department of Computer and Software Engineering, Polytechnique Montreal\\
{\tt\small \{david-alexandre.beaupre, gabilodeau\}@polymtl.ca}}}


%


\maketitle

\begin{abstract}
Multispectral disparity estimation is a difficult task for many reasons: it has all the same challenges as traditional visible-visible disparity estimation (occlusions, repetitive patterns, textureless surfaces), in addition of having very few common visual information between images (\emph{e.g.} color information vs. thermal information). In this paper, we propose a new CNN architecture able to do disparity estimation between images from different spectrum, namely thermal and visible in our case. Our proposed model takes two patches as input and proceeds to do domain feature extraction for each of them. Features from both domains are then merged with two fusion operations, namely correlation and concatenation. These merged vectors are then forwarded to their respective classification heads, which are responsible for classifying the inputs as being same or not. Using two merging operations gives more robustness to our feature extraction process, which leads to more precise disparity estimation. Our method was tested using the publicly available LITIV 2014 and LITIV 2018 datasets, and showed best results when compared to other state of the art methods.
\end{abstract}


%
\IEEEpeerreviewmaketitle

\section{Introduction}
\label{intro}
Disparity estimation from stereo images is one of the fundamental task in the field of computer vision. It can be used to predict the depth in a scene, register images into the same coordinate system and perform object detection. It also has many real-world applications such as for autonomous vehicles and for 3D model reconstruction \cite{7410595, NIPS2015_5644}. 

Many works in the literature focus on disparity estimation in the visible (RGB) domain \cite{ster-zbontar2016stereocnn, ster-luo2016effdl, ster-kendall2017endgeo, ster-chang2018pyramid, Guo_Group_Wise_Correlation_Stereo_Network_CVPR2019}, where both images are captured with RGB cameras. Recently, more works use multispectral image pairs, where one of the images is in the infrared (IR) spectrum \cite{Quan_2019_ICCV, qnet_sensors, 7789530}. This has both advantages and disadvantages. For instance, when we want to detect a given object that has a temperature different from the environment, working with thermal IR images can be beneficial since the desired object will be easily detected. For example, humans usually have a body temperature that is higher or lower than the ambient temperature in public places, making human detection easier. This is especially true if a person's clothes have a low contrast with the environment. For instance, at night time, if someone is walking in a public park, detecting him or her with a thermal IR camera will be easy. However, detecting the same person with a RGB camera would be a lot more challenging. The opposite is also true, where detecting a human in daylight will be easier than in thermal if this human has a body temperature similar to the environment (low thermal contrast). RGB provides as well more information to help describe and identify persons.  Thus, if we work with two cameras from different spectrums, we can get the best of both worlds, since RGB cameras provide more visual information in the case of people, but they require appropriate lighting to do so, while IR camera can capitalize on the thermal contrast between humans and their environment, but will by default remove a lot of details like textures and colors from the image making people identification harder.

\begin{figure}[t] 
	\centering
	\subfigure[] 
    {
  	\includegraphics[width=0.465\linewidth]{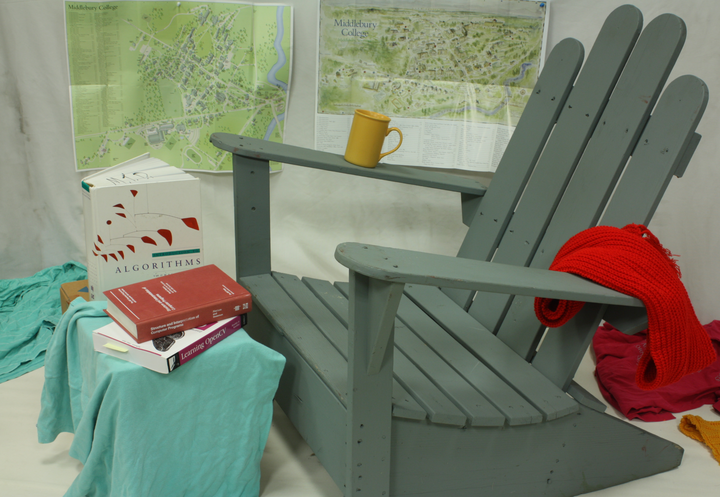}
	}
	\subfigure[] 
    {
  	\includegraphics[width=0.465\linewidth]{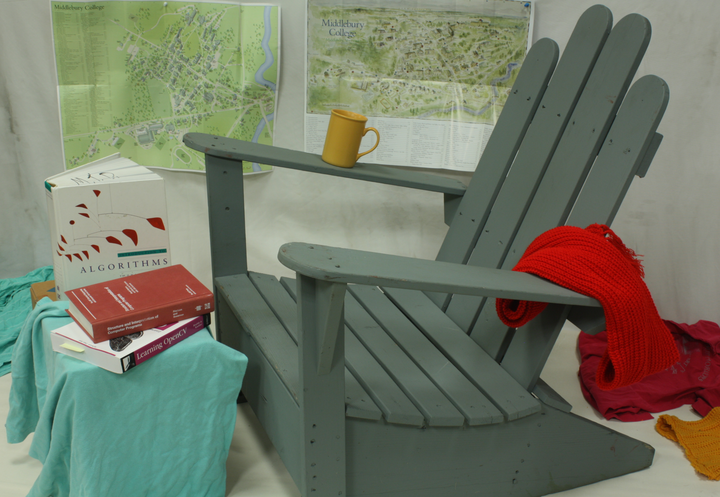} 
	}
	\subfigure[] 
    {
  	\includegraphics[width=0.465\linewidth]{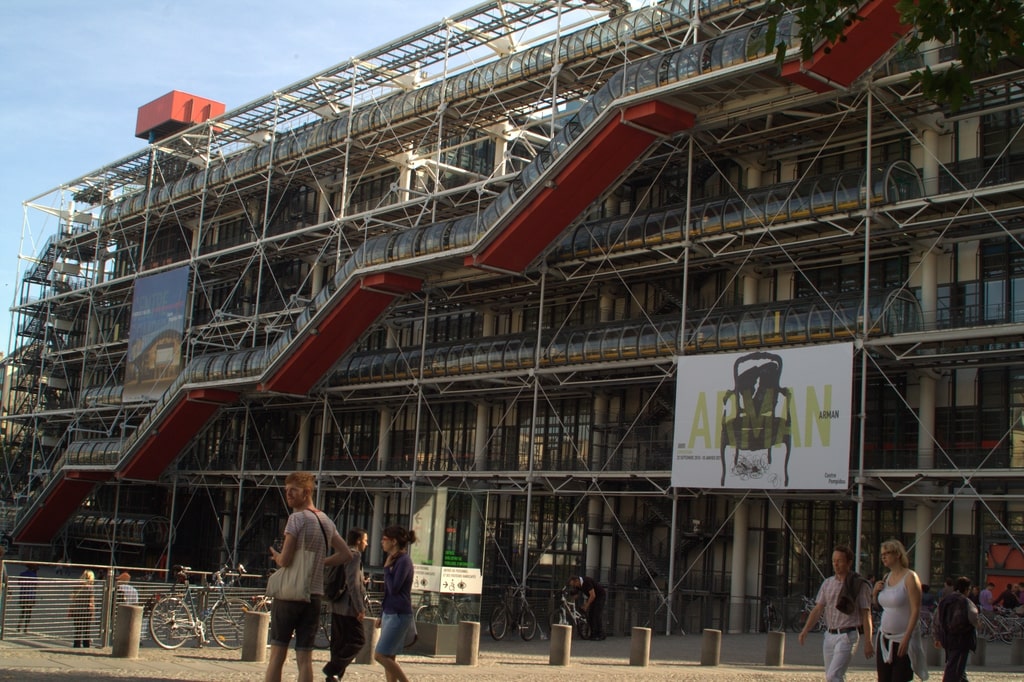}
	}
	\subfigure[] 
    {
  	\includegraphics[width=0.465\linewidth]{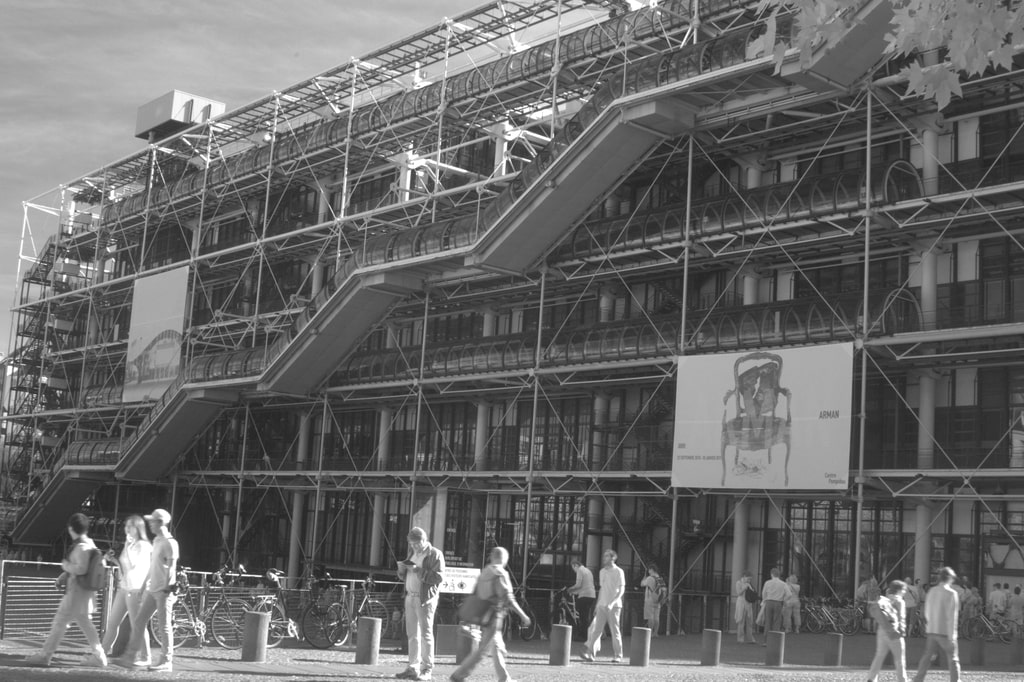} 
	}
	\subfigure[] 
    {
  	\includegraphics[width=0.465\linewidth]{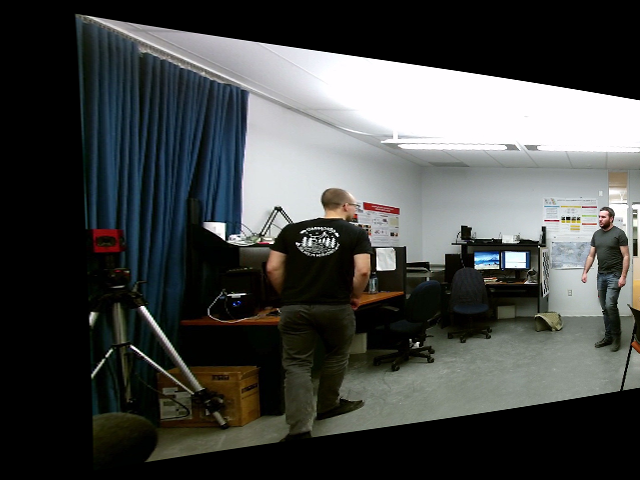}
	}
	\subfigure[] 
    {
  	\includegraphics[width=0.465\linewidth]{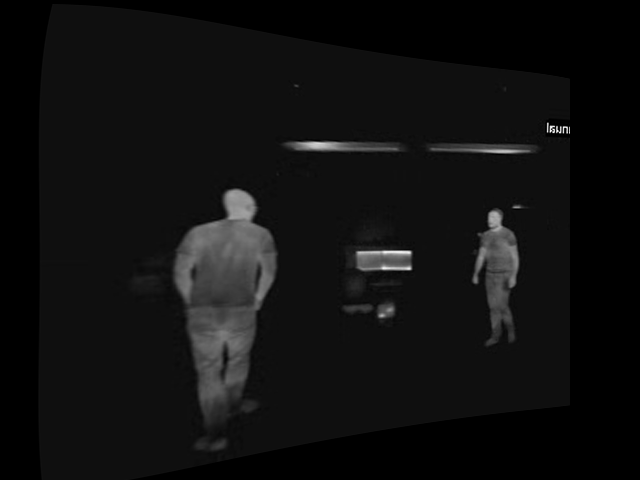} 
	}
  \caption{First row: images from the Middlebury 2014 \cite{middlebury} stereo dataset where we can see a lot of common information between the two images. Second row: images from the VIS-NIR \cite{nir-vis} dataset, which still shows similar textures between both images. Third row: images from the LITIV 2014 \cite{ster-bilodeau2014irvisreg} stereo dataset where the only common information between the two images are the objects emitting heat, with very few common textures.}
\label{fig:diff_multispectral}
\end{figure}

Many works in disparity estimation are designed to work for RGB cameras, where both images have similar content (colors, textures). Lately, several works focus on multispectral images where one image is in RGB while the other one is in the infrared domain. In this case, the amount of similar content in both images will depend on the type of infrared used. However, in all cases, working with IR images reduces the similarity between the two images of the stereo pair, which means that matching the content of said images is harder, and therefore disparity estimation is also harder. 

In figure \ref{fig:diff_multispectral}, we show three different image pairs. It showcases how similarities between image pairs diminish depending on the spectrum of the images. In the case of RGB-RGB images, we can notice that there are a lot of similarities between the two images. Generally, in this domain, the main difficulties are with repetitive patterns, occlusions and textureless areas. However, it is fairly easy to match pixels between two RGB images, as the results on public benchmarks are very good \cite{Geiger_Autonomous_Driving_KITTI_CVPR2012, Menze_Object_Scene_Flow_CVPR2015}. Next, for the case of RGB-NIR (near infrared) images, we observe that there is less common information between the images, but it is still possible to note common objects between both spectrum, as textures are shared and object edges are well defined in both images. 

Lastly, if we take a look at RGB-LWIR (long wavelength infrared or thermal IR) images, we see that there are very few common features between images, mainly the people whom we can match, but every object in the background, if not emitting heat, is very hard to see in the LWIR image. This makes the matching of pixels between both domains much harder than in the previous domain pairs stated above. Our proposed method operates on RGB-LWIR image pairs, which means that we need a method that is able to learn to match features between those domains.

In this paper, we present a new convolutional neural network (CNN) architecture inspired by \cite{Beaupre_2019_CVPR_Workshops, Guo_Group_Wise_Correlation_Stereo_Network_CVPR2019} able to do disparity estimation between RGB-LWIR image pairs. Our model is a domain siamese network, meaning that each image of the stereo pair has its own feature extractor, but both feature extractors have the same components \emph{i.e.} there is no weight sharing between the branches of the siamese network. We are not able to do dense disparity estimation because of the nature of our datasets, which are not densely annotated. That is why we will work with patches instead of images. Our model takes two small square patches as input, and extracts features from those resulting in a feature vector for each image patch. We do two operations on the feature vectors: 1) we do a correlation product between both vectors and forward the result to the correlation head, 2) we do a concatenation between the two vectors and forward it to the concatenation head. Using two different complementary techniques gives more robustness to our network, leading to better performance when compared to using only correlation or concatenation. Both the correlation and concatenation heads consist of fully connected layers outputting the probability of both patches being the same or not. Each classification head has its own loss function, and at testing, we use both classification heads to get the disparity predictions. The source code of our method will be available online at \emph{\url{https://github.com/beaupreda}} upon publication. 

Our paper has the following two main contributions:

\begin{itemize}
    \item We propose a new siamese CNN architecture able to do sparse disparity estimation between multispectral RGB-LWIR image pairs. Our architecture extract features from both image domains and uses both concatenation and correlation to get a probability representing if the input patches are the same or not.
    \item Experiments performed on the LITIV 2014 \cite{ster-bilodeau2014irvisreg} and LITIV 2018 \cite{St-Charles2019} datasets show that our model is able to achieve better performance than all past methods, either using classical descriptors or methods based on CNNs.
\end{itemize}

\section{Related Works}
\label{related}

\begin{figure*}[t]
\begin{center}
   \includegraphics[width=0.9\linewidth]{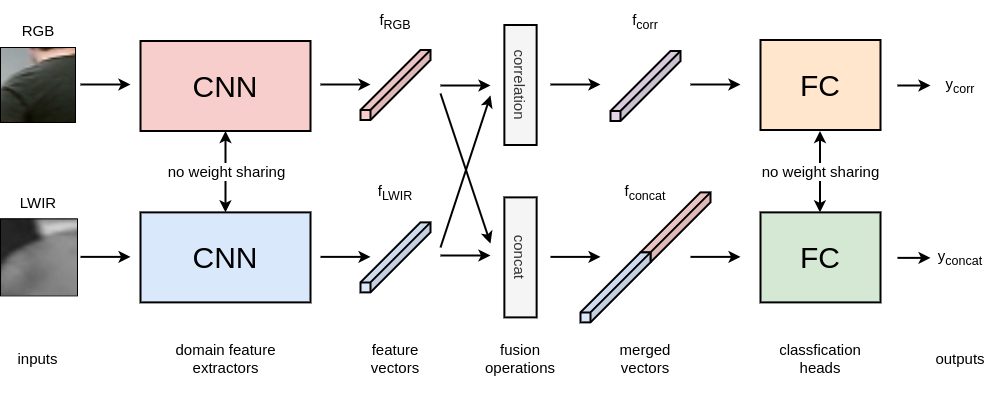}
\end{center}
   \caption{Details of the proposed architecture. We have two CNNs doing domain feature extraction to obtain feature vectors. These vectors are then merged with a correlation and concatenation operation before being passed to the classification heads, which are responsible of classifying the two inputs either as the same or not with a probability score.}
\label{fig:architecture}
\end{figure*}

\subsection{Multispectral Images}
Until recently, the best way to do disparity estimation between the RGB and LWIR domains was with handcrafted feature descriptors, such as SIFT \cite{feat-lowe2004sift} or mutual information \cite{ster-viola1995mi}. Mutual information is an example of a window-based method, meaning that the descriptor is computed as a similarity between pixels inside two candidate windows. The method consists in computing statistics based on the co-occurrences of the pixels inside the windows. This method was able to get the best results among all handcrafted feature descriptors according to a study \cite{ster-bilodeau2014irvisreg}. The sum of squared difference (SSD) \cite{ster-bilodeau2014irvisreg} is another example of window-based method. Other methods like local self-similarity (LSS) \cite{ir-torabi2013lss}, histogram of oriented gradients (HOG) \cite{od-dalal2005hog} and SIFT \cite{feat-lowe2004sift} are based on modeling the distribution of data. LSS is a local descriptor that captures the self-similarity of regions with colors and textures, which makes it the best descriptor in its category \cite{ster-bilodeau2014irvisreg}. SIFT and HOG are both based on gradients and are invariant to illumination changes which is beneficial when working with multispectral images. 

In the last years, there were also works based on CNNs that were designed for patch matching between RGB and IR domains. It is to note that for visible/IR stereo pair datasets, there is not enough data to train CNNs for end-to-end dense stereo matching.  The work of \cite{7789530} gives an interesting comparison of different CNN architectures to do patch matching between the RGB and NIR domains. Aguilera \emph{et al.} \cite{qnet_sensors} propose a quadruplet network inspired by the popular triplet networks \cite{triplet_net}. Quadruplet networks take four input patches, creating two positive pairs and four negatives ones, leading to top performance on the VIS-NIR benchmark \cite{nir-vis}. In the same task, AFD-Net \cite{Quan_2019_ICCV} uses a model based on metric learning using the difference of features between the RGB and NIR images. Zhi \emph{et al.} \cite{ster-zhi2018deepmat} worked with a new RGB-NIR dataset made up of road scenes. It is an unsupervised method that transforms the RGB image into a pseudo-NIR image and uses projection of the pseudo image to do self-supervision. It also uses segmentation to differentiate the materials in the scene. Beaupre \emph{et al.} \cite{Beaupre_2019_CVPR_Workshops} proposed a dual siamese network, effectively working with four inputs to enforce consistency between the predictions of the left and right subnetworks.

\subsection{Stereo Matching}
We will focus on works that use CNNs since they are the most relevant to our work. Zbontar \emph{et al.} \cite{ster-zbontar2016stereocnn} were the first to propose using a CNN to extract features from images instead of using handcrafted feature descriptors. They created a cost volume with the features from the CNN with SGM \cite{Hirschmuller_Stereo_Processing_SGM_MI_PAMI2008} to obtain the disparity map. Luo \emph{et al.} \cite{ster-luo2016effdl} used a correlation product to merge features from the left and right images, resulting in much faster stereo matching than \cite{ster-zbontar2016stereocnn}. Kendall \emph{et al.} \cite{ster-kendall2017endgeo} were the first to propose an end-to-end architecture for stereo matching which is now the default architecture choice for state of the art methods on public benchmarks \cite{Menze_Object_Scene_Flow_CVPR2015}. They used 3D convolutions to learn from the context of features, before doing a regression in order to get the disparity map. Chang \emph{et al.}  \cite{ster-chang2018pyramid} also wanted to improve disparity estimation with help from the context, so they used a spatial pyramid pooling module to extract features at multiple scales and used an hourglass network to regularize the cost volume. The methods of \cite{Yang_SegStereo_Exploiting_Semantic_ECCV2018} and \cite{ster-Xiao_2018_edge} both train a network jointly with another task to improve disparity, namely segmentation and edge detection, respectively. Both papers hypothesize that most errors in disparity estimation come at object borders, so using another task that defines boundary between objects will improve stereo matching. Zhang \emph{et al.} \cite{ster-Zhang_2019_CVPR} focus on the aggregation step by proposing two new layers: semi-global guided aggregation layer (SGA) and local guided aggregation layer (LGA). SGA is basically a version of SGM, but with learnable parameters, while LGA learns to refine thin structures. Guo \emph{et al.} \cite{Guo_Group_Wise_Correlation_Stereo_Network_CVPR2019} focus on the cost volume step by creating two cost volumes: one from the concatenation of features and the other one from the correlation of features. These are then merged in a group-wise manner to form a combine volume of rich features, which leads to better performance than only using concatenation or correlation.

\section{Method}
\label{method}
This section presents an overview of the proposed method, consisting of a domain siamese network with a classification branch and a correlation branch, as well as an explanation of our training and testing methodology, and the implementation details. Figure \ref{fig:architecture} illustrates the global architecture of our model.

\subsection{Network Architecture}
Our network architecture is inspired by previous work in RGB-RGB disparity estimation, as well as RGB-LWIR disparity estimation. It is based on the popular siamese architecture in stereo estimation, where a network takes two inputs and extracts features from these. We modified this base architecture for our needs. Usually, siamese networks share the same feature extractor between the two inputs. However, in our case, we found that having a different feature extractor for each domain gave better results since RBG and LWIR images have different visual appearance. This same conclusion was reached in \cite{multimodal_hybrid_conv} for matching keypoints between the RGB and NIR domains.

\begin{table}[t]
\centering
\caption{Details of our proposed architecture, layer by layer. Layer structure is under the form $k \times k, c$, where $k$ represents the convolutional kernel size and $c$ the number of channels. Output dimension is under the form $h \times w \times c$, $h$ being the height of the patch, $w$ its width and $c$, the number of feature channels. Every convolutional layer is followed by batch normalization \cite{dl-ioffe2015batchnorm} and has a ReLU \cite{dl-nair2010relu} activation function, except for \emph{conv9}. \emph{fc1} and \emph{fc2} also use the ReLU activation, while \emph{fc3} uses a Softmax activation to get probabilities.}
\resizebox{0.8\columnwidth}{!}{%
\begin{tabular}{|c|c|c|}
\hline
Name  & Layer structure & Output dimension \\ \hline
input &                 & $36 \times 36 \times 3$      \\ \hline
\multicolumn{3}{|c|}{CNN}                  \\ \hline
conv1 & $5 \times 5$, $32$       & $32 \times 32 \times 32$     \\ \hline
conv2 & $5 \times 5$, $64$       & $28 \times 28 \times 64$     \\ \hline
conv3 & $5 \times 5$, $64$       & $24 \times 24 \times 64$     \\ \hline
conv4 & $5 \times 5$, $64$       & $20 \times 20 \times 64$     \\ \hline
conv5 & $5 \times 5$, $128$       & $16 \times 16 \times 128$     \\ \hline
conv6 & $5 \times 5$, $128$       & $12 \times 12 \times 128$     \\ \hline
conv7 & $5 \times 5$, $256$       & $8 \times 8 \times 256$     \\ \hline
conv8 & $5 \times 5$, $256$       & $4 \times 4 \times 256$     \\ \hline
conv9 & $4 \times 4$, $256$       & $1 \times 1 \times 256$     \\ \hline
\multicolumn{3}{|c|}{FC}                   \\ \hline
fc1   & $256/512, 128$       & $1 \times 128$     \\ \hline
fc2   & $128, 64$       & $1 \times 64$     \\ \hline
fc3   & $64, 2$       & $1 \times 2$     \\ \hline
\end{tabular}
}
\label{tab:details_arch}
\end{table}

Our architecture takes two square patches as inputs, one from the RGB image and the other from the LWIR image, which will be referred to as $P_{RGB}$ and $P_{LWIR}$ throughout the paper.  Each patch is forwarded into their respective domain feature extractors, one for the RGB features, and the other one for the LWIR features. Both of these CNNs have the exact same structure, which is detailed in table \ref{tab:details_arch}, but each have their own weights. We know from numerous papers that weight sharing between siamese networks is a common practice, specifically in RGB-RGB stereo vision. However, we found that the model performed better when each patch had its own CNN to extract features. We suppose this is because, as mentioned in section \ref{intro}, there is not a lot of common information between images of the RGB-LWIR domains. In the case of images in the RGB domain, sharing weights is logical since we want the network to learn to find similar features from the images. In our multispectral case, since the content of both images can be quite different, we believe it is preferable to extract features separately and then combine them with a fusion operation. The CNNs each produce a 256D feature vector, $f_{RGB}$ and $f_{LWIR}$, representing $P_{RGB}$ and $P_{LWIR}$ respectively. With these feature vectors, we proceed to do two different fusion operations: correlation and concatenation. These two operations are the most common operations in disparity estimation to join features from both images. Each has advantages over the other. Correlation is usually faster to compute and consumes less memory than concatenation, but it loses information \emph{i.e.} it does not keep all the feature information across the channel dimension. On the other hand, concatenation does not lose any information as the whole channel dimension is kept, at the cost of more memory consumption and longer computation time. It was shown that using both a correlation cost volume and a concatenation cost volume improves performance when compared to using only one of the two \cite{Guo_Group_Wise_Correlation_Stereo_Network_CVPR2019}. We take inspiration from this work to use both operations to form the merged feature vectors 

\begin{equation}
    f_{corr} = f_{RGB} \odot f_{LWIR}
\end{equation}

and 

\begin{equation}
    f_{concat} = [f_{RGB}, f_{LWIR}].
\end{equation}

\noindent
$f_{corr}$ is computed as the element-wise product between the feature vectors $f_{RGB}$ and $f_{LWIR}$, so it remains a 256D vector. On the other hand, $f_{concat}$ is a 512D vector, since both $f_{RGB}$ and $f_{LWIR}$ are concatenated. Both merged vectors are then forwarded into their respective classification heads, one for correlation and the other for concatenation. Their structure is detailed in table \ref{tab:details_arch} and both of them outputs a 2D probability vector, representing the probability that the two patches are the same or not. This is therefore a binary classification problem.

\subsection{Training}
The first step in training the network is to extract training patches $P_{RGB}$ and $P_{LWIR}$ around known ground-truth pixels in rectified stereo pairs. $P_{RGB}$ will be centered on pixel $p$ at location $(x, y)$, while $P_{LWIR}$ will be centered on pixel $q$ at location $(x + d, y)$, accounting for the disparity $d$, $x$ and $y$ representing the pixel coordinates in the image space. To make our network more robust, we consider as positive matches, two patches that are positioned from one another at $d\pm 1$ in $x$.

Since we are considering the task as a binary classification problem, we also need samples of negative pairs. In order to create negative pairs, we start by taking all the positive patches located at $(x, y)$ and $(x + d, y)$ and add an offset $o$ to the $x$ values. The range of $o$ is $[-30, -10]$ and $[10, 30]$, meaning that we consider negative samples either to the left or the right of the ground-truth pixels. For every positive pair, we create a negative pair at $(x + o, y)$ and $(x - d + o, y)$ so we have a balanced dataset with as many positive samples as negative ones. The value of $o$ is determined randomly following a uniform distribution for each example. All positive and negative samples are then shuffled together.

We use two binary cross-entropy loss functions to train our network, one for the correlation branch, and the other one for the concatenation branch. They are given by

\begin{equation}
    loss_{corr/concat} = -\frac{1}{N}\sum_{i=1}^{N}gt_{i} log(y_{i}),
    \label{eq:cc}
\end{equation}

\noindent
where $N$ represents the number of samples, $gt_{i}$ the ground-truth value, being either 1 if the patches are the same or 0 if they are not, and $y_i$ the probability of patches being the same or not. The total loss is then given by

\begin{equation}
    loss = loss_{corr} + loss_{concat}.
    \label{eq:total}
\end{equation}

\begin{table*}[]
\caption{Details of the LITIV 2014 data separation into three folds as well as the number of points with data augmentation and from which video these ground-truth points are taken.}
\centering
\begin{tabular}{|c||c|c||c||c|}
\hline
       & \multicolumn{2}{c||}{Training}                       & Validation           & Testing      \\ \hline
       & LITIV 2018                   & LITIV 2014           & LITIV 2014           & LITIV 2014   \\ \hline
Fold 1 & 218 230 (vid04 + vid07 + vid08) & 240 167 (vid02 + vid03) & 35 378 (vid02 + vid03) & 101 433 (vid01) \\ \hline
Fold 2 & 218 230 (vid04 + vid07 + vid08) & 291 720 (vid01 + vid03) & 34 688 (vid01 + vid03) & 76 001 (vid02) \\ \hline
Fold 3 & 218 230 (vid04 + vid07 + vid08) & 320 648 (vid01 + vid02) & 34 220 (vid01 + vid02) & 61 771 (vid03) \\ \hline
\end{tabular}
\label{tab:dataset_sep2014}
\end{table*}

\begin{table*}[]
\caption{Details of the LITIV 2018 data separation into three folds as well as the number of points with data augmentation and from which video these ground-truth points are taken.}
\centering
\begin{tabular}{|c||c|c||c||c|}
\hline
       & \multicolumn{2}{c||}{Training}                       & Validation           & Testing      \\ \hline
       & LITIV 2014                   & LITIV 2018           & LITIV 2018           & LITIV 2018   \\ \hline
Fold 1 & 478 410 (vid01 + vid02 + vid03) & 109 620 (vid07 + vid08) & 44 226 (vid07 + vid08) & 32 192 (vid04) \\ \hline
Fold 2 & 478 410 (vid01 + vid02 + vid03) & 91 904 (vid04 + vid08) & 49 286 (vid04 + vid08) & 38 520 (vid07) \\ \hline
Fold 3 & 478 410 (vid01 + vid02 + vid03) & 99 858 (vid04 + vid07) & 41 566 (vid04 + vid07) & 38 403 (vid08) \\ \hline
\end{tabular}
\label{tab:dataset_sep2018}
\end{table*}

\begin{figure*}[] 
	\centering
	\subfigure[] 
    {
  	\includegraphics[width=0.225\linewidth]{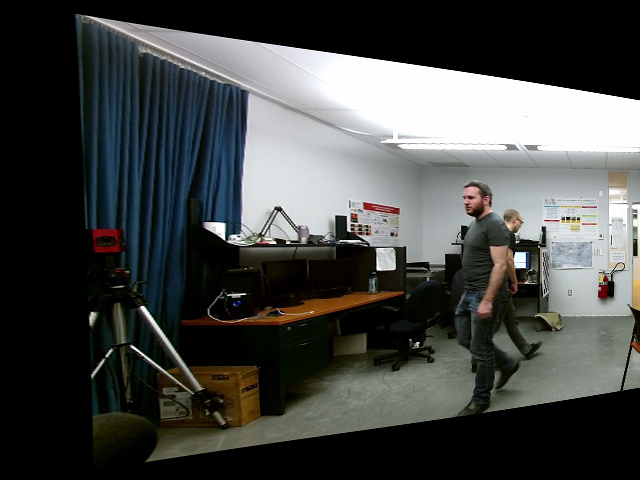}
	}
	\subfigure[] 
    {
  	\includegraphics[width=0.225\linewidth]{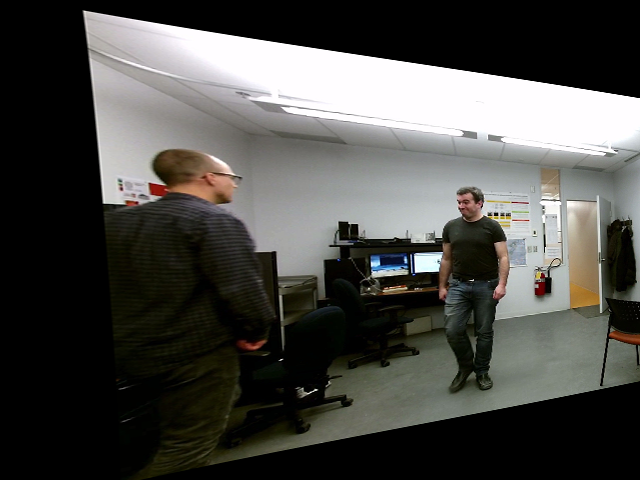} 
	}
	\subfigure[] 
    {
  	\includegraphics[width=0.225\linewidth]{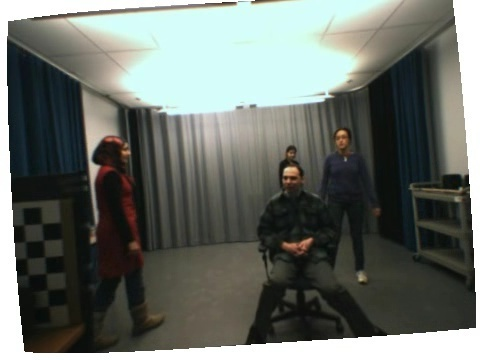}
	}
	\subfigure[] 
    {
  	\includegraphics[width=0.225\linewidth]{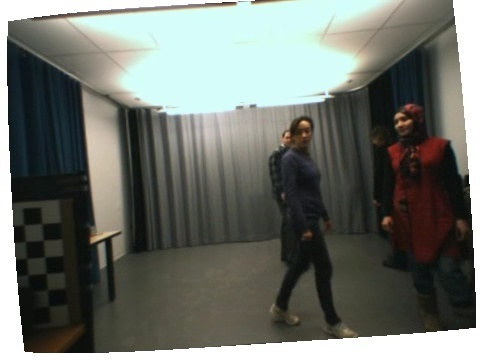} 
	}
	\subfigure[] 
    {
  	\includegraphics[width=0.225\linewidth]{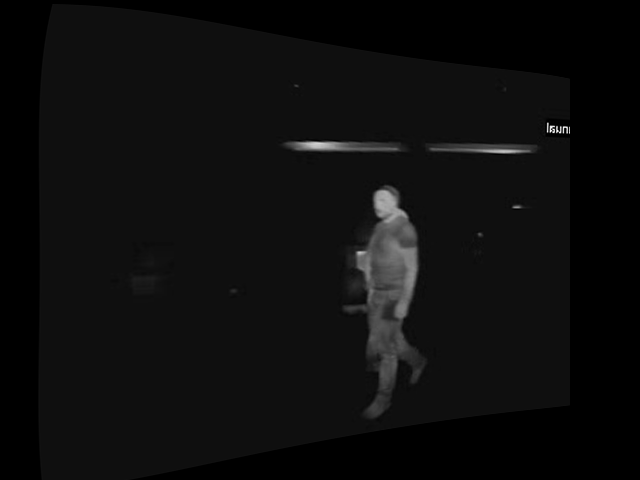}
	}
	\subfigure[] 
    {
  	\includegraphics[width=0.225\linewidth]{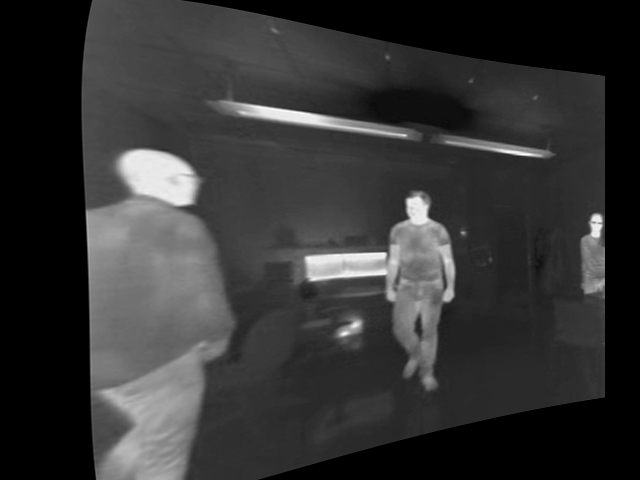} 
	}
	\subfigure[] 
    {
  	\includegraphics[width=0.225\linewidth]{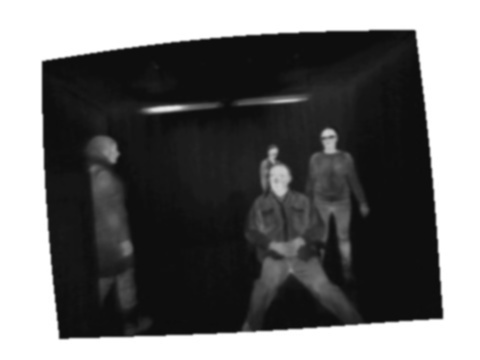}
	}
	\subfigure[] 
    {
  	\includegraphics[width=0.225\linewidth]{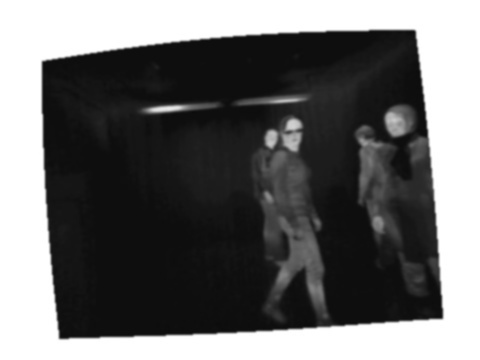} 
	}
  \caption{First two columns: images from the LITIV 2018 \cite{St-Charles2019} dataset. Last two columns: images from the LITIV 2014 \cite{ster-bilodeau2014irvisreg} dataset. These images showcase some of the difficulties present in both datasets. Image pairs (a)-(e) and (d)-(h) show an example of occlusion while pairs (b)-(f) and (c)-(g) show an example of textures not visible in the LWIR domain.}
\label{fig:dataset}
\end{figure*}

\subsection{Prediction}
Since our goal is to predict a disparity, we need to add some operations on top of our network to obtain it. For prediction, we expand the width of $P_{LWIR}$ to take into account the maximum disparity parameter $disp_{max}$ . Half of $disp_{max}$ is added to the width on both side of the center of the original $P_{LWIR}$ patch. $P_{RGB}$ is kept the same size. We forward both patches into the CNNs, leaving us with $f_{RGB}$ a 256D vector and $f_{LWIR}$ a $256 \times disp_{max}$ tensor. Now, for each disparity $d$. we extract the corresponding feature vector from $f_{LWIR}$ leaving us with two vectors, each of 256D. We then forward both vectors into the FCs layers to obtain probabilities of the vectors being the same or not. We do this process for every disparity $d$, and only keep the probability of both patches being the same, which leaves us with a probability vector $p$ of size $disp_{max}$ of the patches being the same at every disparity. We do a disparity regression as in \cite{ster-kendall2017endgeo} \emph{i.e.} a weighted sum of the normalized probabilities multiplied by $d$ to obtain our disparity predictions $\hat{d}_{corr}$ and $\hat{d}_{concat}$ for each head, respectively with

\begin{equation}
    \hat{d}_{corr/concat} = \sum_{d = 0}^{disp_{max}} d \times p_d.
\end{equation}

\noindent
$\hat{d}$, the final disparity prediction, is the mean of the disparity predictions of both branches and is given by

\begin{equation}
    \hat{d} = \frac{d_{corr} + d_{concat}}{2}.
    \label{eq:pred}
\end{equation}

\subsection{Implementation details}
We implemented our network with the PyTorch \cite{pytorch} framework, and as stated earlier, the code will be made publicly available online at \emph{\url{https://github.com/beaupreda}}. The default patch size is $36 \times 36$ and the maximum disparity used for testing is 64. We use the Adam \cite{adam} optimizer to train our network with backpropagation and a starting learning rate $\alpha$ of $0.01$ which is updated to $\frac{\alpha}{2}$ every 40 epochs. We trained our model on every fold of data for 200 epochs, which takes a little more than 12 hours for each fold on a single NVIDIA Titan X.

\section{Experiments}
\label{experiments}
This section will present in details the datasets we used to train and test our model, as well as the data augmentation techniques we used to get more data. It will also show the results we got and discuss them with a comparison against other methods.

\subsection{Datasets}
We use two datasets to train our network and do the evaluation, namely the LITIV 2014 dataset \cite{ster-bilodeau2014irvisreg} and the LITIV 2018 dataset \cite{St-Charles2019}. Some examples of images found in both datasets are shown in figure \ref{fig:dataset}. These images show some of the difficulties of the datasets, like occlusions, where the occluded person is less visible in one of the spectrum, making the matching of patches harder. Another difficulty is the difference of textures between the RGB and the LWIR domains, where a checkered shirt appears as of uniform appearance in the LWIR domain. Another difficulty comes from the number of persons in a given scene, since the number of potential matches increases with the amount of people in a video.

The LITIV 2014 dataset is made of three videos, named vid01, vid02 and vid03, each presenting scenes with people with annotated disparities. The amount of subjects in each video varies from one to five, and they are all walking at different depth in the scene, creating occlusions, which is one of the difficulty in this dataset. The videos respectively have 89, 67 and 53 image pairs for a total of 11 166 points in the first video, 7529 points in the second and 6524 points in the third. The LITIV 2018 is also separated in three videos, named vid04, vid07 and vid08. These videos feature one to three subjects walking in a scene with the additional difficulty of them manipulating some objects emitting heat, such as a kettle. There are 117 images for 4252 ground-truth points in the first video (v04), 144 images for 5653 ground-truth points in the second video (v07) and 89 images for 5277 ground-truth points in the third sequence (v08).

We compared the performance of our model against classical methods that do not need any training phase, so these were tested on all available videos. However, in our case, we need to separate data into three sets, namely training, validation and testing. Since we want to have a fair comparisons against all methods, we split the both datasets into three folds, each fold having a distinct testing set as to have the complete dataset tested. Tables \ref{tab:dataset_sep2014} and \ref{tab:dataset_sep2018} show the separations of the folds with the number of data points in each fold for the LITIV 2014 and LITIV 2018 datasets, respectively. The separation scheme is fairly simple: for a given dataset, we keep one video for testing and use the other two for training and validation. The other dataset is used as training data. For the validation sets, we selected randomly 30 images for LITIV 2014 and 150 images for LITIV 2018.

\subsection{Data Augmentation}

\begin{figure}[t]
\begin{center}
   \includegraphics[width=0.7\linewidth]{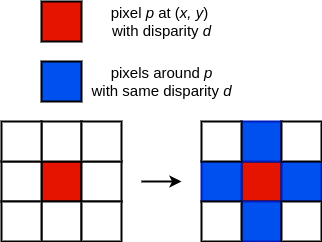}
\end{center}
   \caption{Illustration of our cross data augmentation method. Basically, we add four training points from one pixel from which we have the ground-truth, giving us more training data to reduce overfitting.}
\label{fig:data_aug_cross}
\end{figure}

One problem with RGB-LWIR multispectral datasets is that there is not a lot of available data to train CNNs. For instance, in our case, we use both LITIV datasets, and obtain a little more than 40 000 ground-truth points without any data augmentation. However, with the way we train our network and two data augmentation operations, we are able to effectively increase by a factor of 20 the amount of training data. 

\begin{table*}[t]
\caption{Ablation study on the LITIV 2014 dataset showcasing the difference of performance between our proposed model (last columns) and using only one of the two fusion operations (first and second columns). \textbf{Boldface}: best results}
\centering
\begin{tabular}{|c||c|c|c||c|c|c||c|c|c|}
\hline
       & \multicolumn{3}{c||}{Correlation branch only} & \multicolumn{3}{c||}{Concatenation branch only} & \multicolumn{3}{c|}{Corr + Concat (proposed model)} \\ \hline
       & $ \le1$ px          & $ \le3$ px          & $ \le5$ px         & $ \le1$ px           & $ \le3$ px          & $ \le5$ px           & $ \le1$ px            & $ \le3$ px            & $ \le5$ px            \\ \hline
Fold 1 &     0.524          & 0.859         &   0.984           & 0.551          & 0.894         & 0.981         & \textbf{0.588}           & \textbf{0.901}           & \textbf{0.985}           \\ \hline
Fold 2 &     0.454          & 0.854         &     0.978         & 0.472          & 0.897         & 0.985         &     \textbf{0.474}            & \textbf{0.904}           &    \textbf{0.986}             \\ \hline
Fold 3 &     0.541          & 0.875         &      0.982        & 0.558          & 0.895         & 0.982         & \textbf{0.629}           & \textbf{0.916}           & \textbf{0.989}           \\ \hline
\end{tabular}
\label{tab:ablation}
\end{table*}

\begin{table*}[t]
\centering
\caption{Comparison of our model against two other methods on the LITIV 2018 dataset. \textbf{Boldface}: best result.}
\begin{tabular}{|c||c|c||c|c||c|c||c|c|}
\hline
                             & \multicolumn{2}{c||}{Fold 1} & \multicolumn{2}{c||}{Fold 2} & \multicolumn{2}{c||}{Fold 3} & \multicolumn{2}{c|}{Overall} \\ \hline
                             & $ \le$ 1 px         & $ \le$ 4 px         & $ \le$ 1 px         & $ \le$ 4 px         & $ \le$ 1 px         & $ \le$ 4 px         & $ \le$ 1 px          & $ \le$ 4 px         \\ \hline
DASC Sliding Window \cite{St-Charles2019}         & 0.104        & 0.265        & 0.086        & 0.236        & 0.121        & 0.289        & 0.104         & 0.263        \\ \hline
Multispectral Cosegmentation \cite{St-Charles2019} & 0.253        & 0.562        & 0.236        & 0.531        & 0.307        & 0.678        & 0.265         & 0.590        \\ \hline
Proposed Model               & \textbf{0.480}        &\textbf{ 0.943}        & \textbf{0.446}        & \textbf{0.877}       & \textbf{0.406}        & \textbf{0.972}        & \textbf{0.442}         & \textbf{0.930}        \\ \hline
\end{tabular}
\label{tab:results2018}
\end{table*}

\begin{table}[t]
\centering
\caption{Comparison of our model against other methods on the LITIV 2014 dataset. Patch sizes are in parentheses. \textbf{Boldface}: best result, \textit{italic}: second best.}
\begin{tabular}{|c|c|}
\hline
Method                     & $\le 3$ px \\ \hline
Proposed Model ($36 \times 36$) & \textbf{0.906} \\ \hline
Siamese CNNs \cite{Beaupre_2019_CVPR_Workshops} ($37 \times 37$)            & 0.779  \\ \hline
Mutual Information \cite{ster-bilodeau2014irvisreg} ($40 \times 130$)        & \textit{0.833}  \\ \hline
Mutual Information \cite{ster-bilodeau2014irvisreg} ($20 \times 130$)        & 0.775  \\ \hline
Mutual Information \cite{ster-bilodeau2014irvisreg} ($10 \times 130$)        & 0.649  \\ \hline
Fast Retina Keypoint \cite{ster-bilodeau2014irvisreg} ($40 \times 130$)      & 0.641  \\ \hline
Local Self-Similarity \cite{ster-bilodeau2014irvisreg} ($40 \times 130$)     & 0.734  \\ \hline
Sum of Squared Differences \cite{ster-bilodeau2014irvisreg} ($40 \times 130$) & 0.656  \\ \hline
\end{tabular}
\label{tab:results}
\end{table}

The first data augmentation operation that we did is what we call the cross duplication. This process is illustrated in figure \ref{fig:data_aug_cross} and basically consists in giving the same disparity to neighbors of a given ground-truth pixel. If we have a ground-truth pixel $p$ at location $(x, y)$ with disparity $d$, we simply create four new ground-truth points with the same disparity $d$ at locations $(x \pm 1, y \pm 1)$, giving the shape of a cross to our ground-truth points. We remove any duplicates that this process can create, as two ground-truth point having a common neighbor will duplicate a new point. This operation basically adds a factor of five to the number of original disparity points. One important thing to note, however, is that from the original 40 000 points, some of them cannot be used in this step since they are too close to the image border, and we cannot extract patches around those points. Since all our points are on people, this does not happen very often, mostly when a subject enters the scene.

The second data augmentation operation we performed is the mirroring of images \emph{i.e.} flipping the images around the $y$ axis. This operation doubles the amount of data points from the vanilla data. Now, if we combine both data augmentation technique and use the fact that we have to create negative samples for each ground-truth point that we have (doubles the number of data), these operations give us a factor of 20 when compared to the original training data. With this amount of training data, we reduce the chance of overfitting and increase the robustness of our network.

\subsection{Results}
The performance measure we use to compare our model to other methods is the recall, which computes the number of predictions $\hat{d}$ that our network made that is at a distance of $t$ pixels or less from the ground-truth. We formally define it by 

\begin{equation}
    recall = \frac{1}{N} \sum_{i = 1}^{N} |\hat{d}_i - gt_i| \le t,
    \label{metric}
\end{equation}

\noindent
where $N$ stands for the number of samples, $\hat{d}_i$, the disparity prediction for the $i$th example and $gt_i$ the ground-truth of the $i$th example. This evaluation methodology is the same as the other methods that we compare ourselves to, so we can directly report the results from those papers.

Table \ref{tab:ablation} presents the results from our ablation study which evaluates the performance of our model when only one fusion operation is considered. For the correlation branch only results, we trained our network without the concatenation branch. The same process was applied to the concatenation only results, where only the concatenation branch was considered during training. The last column is our proposed model with both branches present during training, and the final disparity prediction $\hat{d}$ is the average of the predictions of both heads. The most important result from this table is that using both branches leads to better results for disparity prediction. This is expected since using both correlation and concatenation leads to better extracted features by the CNNs, and therefore better predictions. We also observe that by itself, the concatenation branch has better results than the correlation branch. This result is logical since the concatenation operation keeps more features (512D feature vector) compared to the correlation operation (256 feature vector). Also, it is the same conclusion reached in RGB-RGB disparity estimation where networks who build a concatenation cost volume have better performance than the ones who use a correlation operation.

Table \ref{tab:results} shows the results of our method compared to a CNN-based method from \cite{Beaupre_2019_CVPR_Workshops} and several methods based on handcrafted feature descriptors, as reported in \cite{ster-bilodeau2014irvisreg}. The recall obtained by our proposed model is the weighted average (by the number of test points) of the recalls obtained for the three separate folds reported in table \ref{tab:ablation}. We can see that our proposed method surpasses every other method by a large margin. We improve the past results based on deep learning by a little bit more than $0.12$, while also having an improvement of around $0.07$ over mutual information \cite{ster-viola1995mi}, which was the best on the LITIV 2014 dataset. We also achieve this performance while working with less pixels than any other methods. Our performance is significantly superior to siamese CNNs with around the same number of pixels to make our decision, but if we compare our small $36 \times 36$ patches to the other methods, we can notice that they use up to four times more pixel, and yet, we are able to obtain better disparity predictions.

Table \ref{tab:results2018} presents the results on the LITIV 2018 dataset against two approaches, one being classical, while the other is a sophisticated cosegmentation method applying belief propagation to optimize disparity estimation and segmentation jointly. Here, we can see that our method is superior two the other two approaches, having an overall better performance of more than $0.20$ for the smallest threshold, and more than $0.30$ for the bigger one.

We believe this demonstrates that our method is robust and that CNNs-based methods are able to be competitive on hard tasks like disparity estimation between RGB-LWIR image pairs. We believe that these performances are due to mainly two features of our model: first, the domain feature extractors, which are responsible to get features from both domains separately and the use of two fusion operations which forces the network to learn better representations of the multispectral images.

\section{Conclusion}
\label{conclusion}
In this paper, we presented a new model able to do sparse multispectral disparity estimation between image pairs from the RGB and LWIR domains. Our model uses a siamese domain feature extractor, which extracts features independently for both images, as opposed to traditional siamese networks who shares the same weights for both images. We believe that it is preferable to keep the feature extractions separate in the case of images from different spectral domains. We use two operations to merge the features extracted by our CNNs, namely correlation and concatenation. Using the two jointly is shown to improve the performance of our network. We believe that by using the two fusion operations, we augment the learning capability of our network and make it more robust which leads to better performance. Evaluation on public datasets shows that our method is significantly better then all the other methods tested for sparse multispectral disparity estimation.


\section*{Acknowledgment}
This work was made possible by a scholarship from the National Sciences and Engineering Research Council of Canada (NSERC). We would also like to thank NVIDIA Corporation for their donation of a Titan X GPU used for this research.




%

\bibliographystyle{IEEEtran}
\bibliography{egbib}




\end{document}